# Analysis of Truck Driver Behavior to Design Different Lane Change Styles in Automated Driving

Zheng Wang, Muhua Guan, Jin Lan, Bo Yang, Tsutomu Kaizuka, Junichi Taki,
and Kimihiko Nakano

*Abstract*—Lane change is a very demanding driving task and number of traffic accidents are induced by mistaken maneuvers. An automated lane change system has the potential to reduce driver workload and to improve driving safety. One challenge is how to improve driver acceptance on the automated system. From the viewpoint of human factors, an automated system with different styles would improve user acceptance as the drivers can adapt the style to different driving situations. This paper proposes a method to design different lane change styles in automated driving by analysis and modeling of truck driver behavior. A truck driving simulator experiment with 12 participants was conducted to identify the driver model parameters and three lane change styles were classified as the aggressive, medium, and conservative ones. The proposed automated lane change system was evaluated by another truck driving simulator experiment with the same 12 participants. Moreover, the effect of different driving styles on driver experience and acceptance was evaluated. The evaluation results demonstrate that the different lane change styles could be distinguished by the drivers; meanwhile, the three styles were overall evaluated as acceptable on safety issues and reliable by the human drivers. This study provides insight into designing the automated driving system with different driving styles and the findings can be applied to commercial automated trucks.

*Index Terms*— Intelligent transportation systems, intelligent vehicles, automated driving, human factors, man-machine systems, user-centered design;

## I. Introduction

AUTOMATED driving nowadays has drawn much attention because it has great potential to reduce driver workload and to improve driving safety. Due to the labor shortage and aging problem of truck drivers, there is an even more urgent need for automated driving of trucks [1, 2]. Lane change is a very demanding driving task that induces tremendous numbers of traffic accidents [3], so the development of automated lane change systems is arousing wide interest, which has been reviewed by [4, 5].

According to SAE "Levels of Driving Automation" Standard [6], before fully autonomous driving is available in the market, drivers are stilled required to monitor the driving environment and to take over the driving task when necessary on Automation Levels 2 and 3. Considering this, it is important to make the driver feel safe and comfortable with the automated driving system so that the system will be accepted from the viewpoint of human factors [7]. Thus, the problem here is how to improve driver acceptance on the automated driving system.

Human-centered automation has been proved to reduce the likelihood of human-machine misunderstanding and to improve cooperation [8]. An automated driving system designed for an average driver would be conservative to some drivers considering safety issues, so the personalized driver model has been developed and applied to driver-automation systems [9, 10]. A variety of driver model has been developed as a result of the specific requirements and different applications [11-13]. In recent years, benefiting from the growing amount of available data, plenty of data-driven methods have been developed to design human-like automated driving systems [14-18]. In the work carried out by [15], driving behaviors were extracted from expert human drivers using the deep learning method, which was applied to an autonomous vehicle to successfully reproduce proactive driving behaviors. Another commonly used data-driven method, support vector machine, was developed to capture the lane change decision behavior of human drivers, and was integrated into a model predictive control framework to create a personalized automated driving system [17]. It has been found that human-like automated driving systems can increase the driver's sense of agency which makes the driver feel more involved in the driving task, so the driving experience and driver acceptance of the automated driving system are improved [19]. In addition, a higher sense of agency could reduce driver distraction and inattention during automated driving [20], and could shorten the response time for a take-over request in a critical event [21].

In order to improve driver acceptance on the automated driving system, adaptable automation has been proposed, which

This work was mainly supported by Hino Motors, Ltd. and partially supported by a Grant-in-Aid for Early-Career Scientists (no. 19K20318) from the Japan Society for the Promotion of Science.

Corresponding author: Zheng Wang (e-mail: z-wang@iis.u-tokyo.ac.jp).

Z. Wang, M. Guan, J. Lan, B. Yang, T. Kaizuka, and K. Nakano are with the Institute of Industrial Science, The University of Tokyo, Tokyo, 153-8505 Japan

Junichi Taki is with the Hino Motors, Ltd., Tokyo, 205-8660

plays an important role in the human-centered approach for automation system design [8, 22, 23]. Under different driving situations, drivers prefer to choose a different type or level of automation for achieving a better performance [22, 24]. Considering that individual drivers have different driving styles, aggressive drivers would want aggressive driving systems and conservative drivers would want conservative driving systems [9]. Even for the same driver, a more aggressive driving style would be preferred when he/she is in a hurry time. To this end, the users/drivers should be able to adjust the driving style of automated vehicles to their preference to improve their driving comfort [25]. In terms of the lane change decision-making phase, a time-efficient recognition method was used to automatically label the decision-making data into three styles, including moderate, vague, and aggressive, and the recognition accuracy and stability were verified [26]. In terms of the lane change maneuver phase, a dynamic model was proposed to reflect the driver control strategies of adjusting longitudinal and latitudinal acceleration with different driving styles, including slow and careful as well as sudden and aggressive [27]. Although these properly designed systems can provide different driving styles, the problem of evaluating their effect on driving experience, e.g. from the viewpoint of ego and surrounding vehicles, was not further addressed by these studies.

On the other hand, the above-mentioned previous research has mainly focused on passenger cars and limited attention has been paid to trucks. The necessity of developing automated driving systems for trucks is certain by considering the high workload of truck drivers and the economic costs [28]. The vehicle dynamics of a truck are much different from a passenger car resulting from the truck's length, size, weight, and maneuverability. Moreover, truck drivers are usually well trained and they are professional to make a quick and smooth lane change in most situations [29].

The aim of this study was to design and evaluate an automated lane change system with different driving styles based on analysis and modeling of truck driver behavior. The hypothesis was that the different driving styles, including aggressive, medium, and conservative, can be distinguished by drivers from viewpoint of ego and surrounding vehicles; meanwhile, the driver acceptance of the system reliability can be demonstrated. It is expected that the findings in this study will be applied to commercial automated trucks.

This paper is organized as follows. Section II introduces an automated lane change model of truck drivers. Section III presents the designing of an automated lane change system with different driving styles, which uses the driver model parameters identified from a driving simulator experiment. In Section IV, the evaluation of the designed automated lane change system by conducting another driving simulator experiment is presented, and the evaluation results are illustrated. Discussions on the system design and evaluation are addressed in Section V, and conclusions and future work are described in Section VI.

## II. MODEL OF AUTOMATED LANE CHANGE

In this section, a discrete lane change model based on gap acceptance is presented. There are three phases of the lane change process, including the car following phase, lane change decision-making phase, and dynamics control phase during the lane change maneuver.

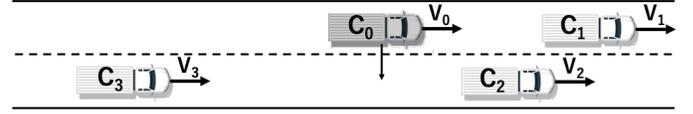

Fig. 1. Lane change scenario.

### A. Related Work

Lane change tasks are demanding and require the information of surrounding vehicles. Previous research has classified the lane change task of trucks as two subcategories: mandatory lane change and discretionary lane change [30]. This paper addresses the discretionary lane change with a discrete decision model based on gap acceptance [31, 32].

### B. Lane Change Scenario

The lane change scenario is shown in Figure 1, in which four trucks run on two adjacent lanes. At a certain moment, the vehicle $C_1$, which runs in front of the ego vehicle $C_0$, begins to decelerate, while $C_2$ and $C_3$ keep running at a constant speed. After the deceleration, $C_1$ keeps running at a relatively lower speed compared to $C_2$ and $C_3$. In order to maintain a higher driving speed, $C_0$ attempts to make a lane change by detecting the distance to the surrounding vehicles. If the gap acceptance to surrounding vehicles is satisfied, $C_0$ will make a lane change to enter between $C_2$ and $C_3$; otherwise, it will continue to perform the car following task behind $C_1$.

During the car following phase, the ego vehicle $C_0$ detects the distance to surrounding vehicles $C_1$, $C_2$, and $C_3$, indicated by $X_1$, $X_2$, and $X_3$, respectively, as shown in Figure 2 (a). In addition, the speed of each vehicle is indicated by $V_0$, $V_1$, $V_2$, and $V_3$, respectively.

At the timing $t_1$, $C_0$ starts to make a lane change maneuver. At the timing $t_2$, the center of gravity of $C_0$ reaches the lane boundary, as shown in Figure 2 (b). At this timing, the distance from $C_0$ to $C_1$ is marked as $D_1$. At the timing $t_3$, the lane change maneuver is completed, when $C_0$ reaches the target lane and its yaw angle becomes 0 degrees, as shown in Figure 2 (c). At this timing, the distances from $C_0$ to $C_2$ and from $C_0$ to $C_3$ are marked as $D_2$ and $D_3$, respectively.

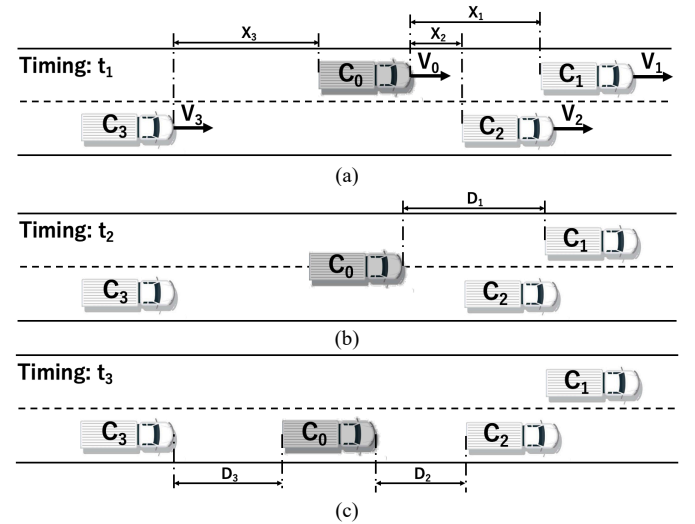

Fig. 2. Lane change process with three stages: (a) Starting point of lane change at timing $t_1$, (b) Crossing lane point of lane change at timing $t_2$, and (c) Ending point of lane change at timing $t_3$.

*C. Car Following Phase*

The ego vehicle $C_0$ keeps following the lead vehicle $C_1$ until a lane change decision is made and a maneuver is performed. If the gap acceptance is not satisfied, $C_0$ continues to perform the car following task. In this paper, an adaptive cruise control based car following model, which automatically adjusts the vehicle speed to maintain a safe distance from the lead vehicle, is presented, as shown in Figure 3. In the car following model, it is needed to make sure that $D_1$ is always larger than $S_1$ ($S_1$ is the threshold of safe distance for $D_1$); thereby, whenever the gap acceptance of $D_2$ and $D_3$ is satisfied, the ego vehicle can make a lane change. In addition, when $V_0$ is greater than $V_1$, the distance from $C_0$ to $C_1$ gradually reduces. If $C_0$ starts to decelerate when $D_1$ is equal to $S_1$, $D_1$ will become lower than $S_1$. Thus, the deceleration process of $C_0$ with the variation of distance to $C_1$ is also considered. The relation between $D_1$ and $S_1$ is indicated by the following equation,

$$D_1 > \frac{V_0^2 - V_1^2}{2a_d} - V_1 \frac{V_0 - V_1}{a_d} + S_1 \quad (1)$$

If the gap acceptance to surrounding vehicles is satisfied, which will be described in the next part, $C_0$ stops following $C_1$ and makes a lane change.

*D. Lane Change Decision based on Gap Acceptance*

The thresholds of safe distances for $D_1$, $D_2$, and $D_3$ are named as $S_1$, $S_2$, and $S_3$, respectively. If $D_1>S_1$, $D_2>S_2$, and $D_3>S_3$ are all satisfied, $C_0$ will make a lane change. The flowchart of the lane change decision is shown in Figure 4.

$D_1$, $D_2$, and $D_3$ are calculated by predicting the relative distances to surrounding vehicles during the lane change maneuver, which are expressed as

$$D_1 = X_1 + V_1(t_2 - t_1) - (V_0(t_2 - t_1) + \frac{1}{2}a_x(t_2 - t_1)^2) \quad (2)$$

$$D_2 = X_2 + V_2(t_3 - t_1) - (V_0(t_3 - t_1) + \frac{1}{2}a_x(t_3 - t_1)^2) \quad (3)$$

$$D_3 = X_3 - V_3(t_3 - t_1) + (V_0(t_3 - t_1) + \frac{1}{2}a_x(t_3 - t_1)^2) \quad (4)$$

where $a_x$ is the constant longitudinal acceleration during the lane change maneuver, which will be explained in the next part.

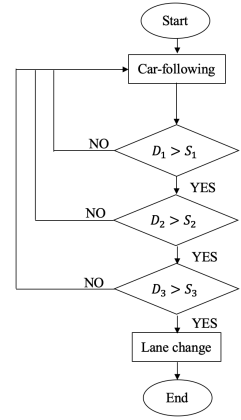

Fig. 4. Flowchart of lane change model.

*E. Dynamics Control Phase During Lane Change Maneuver*

Longitudinal and lateral accelerations are two important parameters for the dynamics control phase during the lane change maneuver. It is assumed that the longitudinal acceleration $a_x$ is a constant value during the lane change maneuver. The lateral acceleration $a_{y12}$ is a constant value during the first half stage of lane change maneuver (from timing $t_1$ to $t_2$) and $a_{y23}$ is also a constant value during the second half stage of lane change maneuver (from timing $t_2$ to $t_3$). They are expressed as

$$a_x = \frac{v_{t3} - v_{t1}}{t_3 - t_1} \quad (5)$$

$$a_{y12} = \frac{2l_1}{(t_2 - t_1)^2} \quad (6)$$

$$a_{y23} = \frac{2l_2}{(t_3 - t_2)^2} \quad (7)$$

where $v_{t1}$ and $v_{t3}$ are longitudinal speed of $C_0$ at timing $t_1$ and $t_3$, respectively, and $l_1$ and $l_2$ are the lateral distance variation of $C_0$ from timing $t_1$ to $t_2$ and from timing $t_2$ to $t_3$, respectively.

### III. Design of Automated Lane Change System

A truck driving simulator experiment was conducted to collect the driving data, which were used to identify the parameters of the proposed lane change model. The identified parameters were classified into three types, namely aggressive, medium, and conservative, to represent different driving styles.

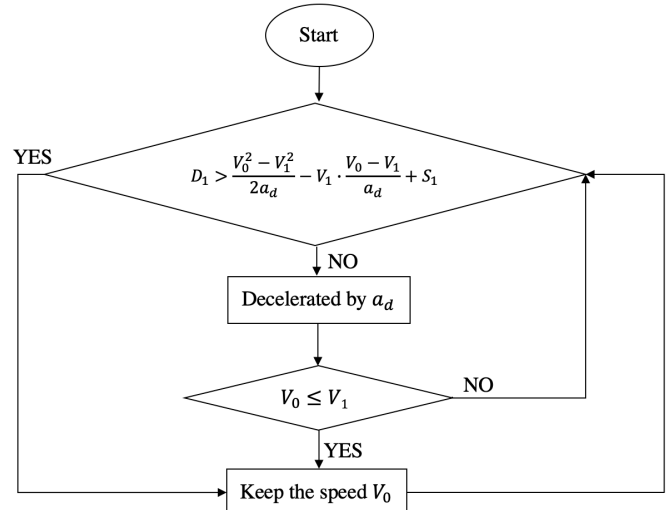

Fig. 3. Flowchart of car following model.

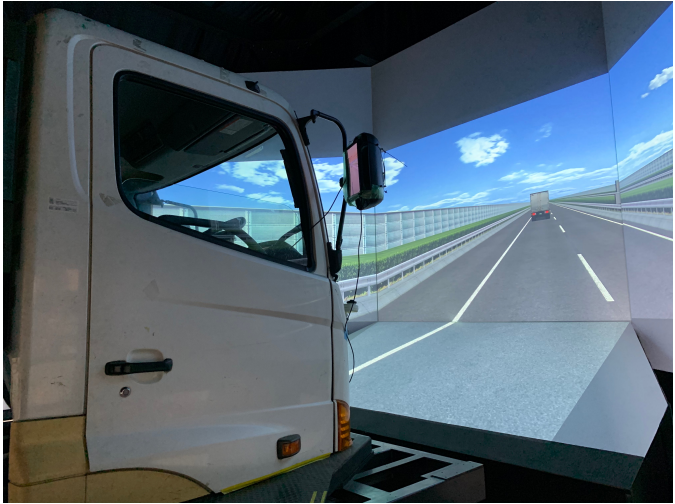

Fig. 5. Truck driving simulator with right rear-view mirror.

## A. Driving Simulator Experiment

### 1) Participants

Twelve professional truck drivers (11 males and 1 female) participated in the experiment. Their age ranged from 26 to 60 years (mean = 42.9, SD = 8.2). All the participants had valid Japanese driver's licenses for heavy vehicles. Their driving experience ranged from 2 to 26 years (mean = 14.6, SD = 8.2). The experiment received ethical approval from the Ethics Committee of Interfaculty Initiative in Information Studies, The University of Tokyo.

### 2) Apparatus

The experiment was conducted in a truck driving simulator, which consisted of a real truck cabin, two rear-view mirrors, and a 140° field-of-view driving scene visualized by four projectors, as shown in Figure 5. The two rear-view mirrors were emulated by using two monitors for presenting the rear environment. The visual scene was updated at a rate of 120 Hz. The truck driving simulator also had a 6 degree-of-freedom motion platform, which was capable to provide real driving sensations to the drivers including the feeling of accelerating, braking, turning, and vibration caused by rough roads. Within the cabin, the cockpit of a real truck was mimicked by providing an automatic transmission, brake pedal, accelerator pedal, shift handle, actuated steering wheel, and dashboard. Raw data of driving performance were recorded in the host computer of the driving simulator at a sampling rate of 120 Hz.

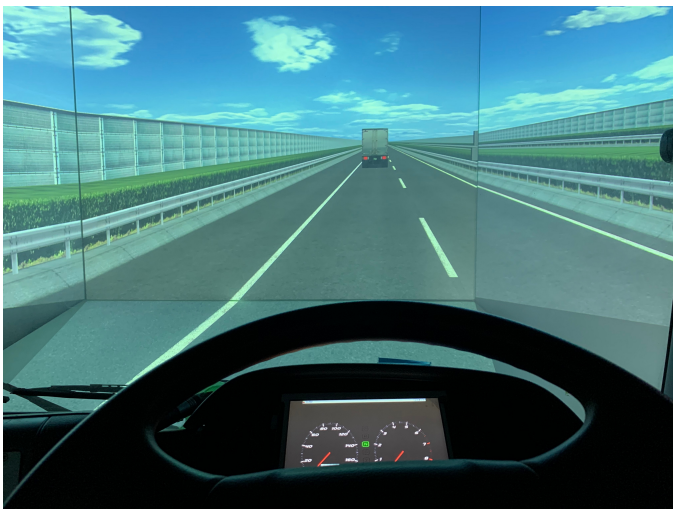

Fig. 6. Front driving scene in the experiment.

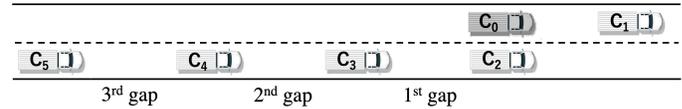

Fig. 7. Lane change scenario in experiment for system design.

### 3) Scenario

The driving environment was a two-lane highway road with an emergency lane on the left, as shown in Figure 6. The driving scenario of the lane change task is shown in Figure 7. In total, there were six trucks, and the body length of each truck was 12 m. In the left lane of the highway, there were two vehicles including a lead vehicle (corresponding to $C_1$ in Figure 1) and an ego vehicle (corresponding to $C_0$ in Figure 1). In the right lane of the highway, there were four trucks with different gap distances. The gap distance was defined as the distance between the rear end of the leading vehicle and the front bumper of the following vehicle.

At the beginning of the driving task, the vehicles accelerated to a speed of 80 km/h and kept the speed. After a certain period, the vehicle $C_1$ began to decelerate with a constant deceleration rate of 5 m/s$^2$ until reaching a speed of 70 km/h, and then kept the speed. Other vehicles in the right lane continued to run at a constant speed of 80 km/h. When $C_1$ slowed down, the ego vehicle $C_0$ also needed to decelerate to keep a safe distance to $C_1$. In the meantime, $C_0$ attempted to make a lane change by detecting the distance to the surrounding vehicles. The participants in $C_0$ were asked to make a lane change in their natural driving way. The driving task ended after the lane change was completed.

In order to determine the gap distance setting among the vehicles in the right lane, a pilot experiment was conducted. According to the pilot experiment results, a gap distance of 40 m was considered to be short for the professional driver to make a lane change; a gap distance of 70 m was considered to be long enough for making a lane change. Therefore, in the formal experiment, four vehicles ran in the right lane with three gap distances of 50, 60, and 70 m. These three gap distances appeared in the form of permutations and combinations with six experimental conditions, as shown in Table I. In addition, the distance between $C_2$ in the right lane and $C_1$ in the left lane was set as 55 m. During the driving task, the participants were asked to follow the lead vehicle $C_1$ and not to fall behind the vehicle $C_2$, so that the first gap distance between $C_2$ and $C_3$ can be taken into consideration for the lane change decision.

TABLE I
CONDITIONS OF EXPERIMENT FOR SYSTEM DESIGN

|  | Cond. 1 | Cond. 2 | Cond. 3 | Cond. 4 | Cond. 5 | Cond. 6 |
| --- | --- | --- | --- | --- | --- | --- |
| 1st gap (m) | 50 | 50 | 60 | 60 | 70 | 70 |
| 2nd gap (m) | 60 | 70 | 50 | 70 | 50 | 60 |
| 3rd gap (m) | 70 | 60 | 70 | 50 | 60 | 50 |

Each condition was repeated twice, and therefore, each participant drove a total of 12 trials (including 6 trials with different conditions and then the repeated 6 trials). The order of the experimental conditions presented to the participants was partially counterbalanced by using a Latin Square [33].

*4) Procedure*

On the experiment day, first, the participants were asked to sign a consent form after the experiment contents were explained. The participants then filled out a questionnaire about their personal information and driving experience. After that, they got into the driving simulator and adjusted their seat to achieve a normal driving position. Participants were asked to follow Japanese traffic rules and to drive as naturally as possible. The participants drove two practice trials of the driving task in order to familiarize themselves with the driving simulator and the lane change task.

After the practice trials, the participants were allowed to rest for 5 min. After the rest, the formal experimental session started, in which the participants drove 12 trials. The participants were instructed about the three gaps in the right lane at the beginning of the formal session; however, the participants were not informed about the order that the three gaps occurred. The entire experiment took approximately 120 min per participant.

*5) Measurement*

The relative distance of the ego vehicle to the surrounding vehicles and the speed of the vehicles during the driving task were measured. The longitudinal acceleration and lane change duration of the ego vehicle were also measured.

*B. Identified Parameters*

The parameters of the proposed lane change model (described in Section II) were identified from the measured data, including relative distances to surrounding vehicles, longitudinal acceleration, and lane change durations. The distribution of the parameter values, including $D_1$, $D_2$, $D_3$, $(t_2-t_1)$, $(t_3-t_1)$, and $a_x$ are shown in Figure 8.

The identified parameters were used to design the automated lane change system and the designed system was evaluated by another driving simulation experiment described in Section IV. It was assumed that the lateral acceleration during the lane change maneuver was constant, as shown in equations (6) and (7). $a_{y12}$ represents the lateral acceleration from timing $t_1$ to $t_2$ and $a_{y23}$ represents the lateral acceleration from timing $t_2$ to $t_3$. Through a trial-and-error process, we determined the steering wheel angles as inputs to control the vehicle during the automated lane change maneuver which fitted the desired lateral acceleration.

*C. Classification of Driving Styles*

Three driving styles of lane change were considered in this study including aggressive, medium, and conservative ones. They were classified by choosing the 25th, 50th, and 75th percentile of the identified parameters of relative distances to surrounding vehicles (including $D_1$, $D_2$, and $D_3$), respectively. The hypothesis was that choosing 25th, 50th, and 75th percentile can make the different driving styles distinguishable by drivers, and meanwhile, the three styles can be evaluated as safe and reliable by drivers. On the one hand, choosing the gaps with more percentile (e.g. 10th, 50th, and 90th) between different styles could make it easier for drivers to distinguish them, but it also could lead to unsafe and unreliable feelings, especially in the case of aggressive driving style. On the other hand, choosing the gaps with less percentile could make it harder for drivers to distinguish the different driving styles.

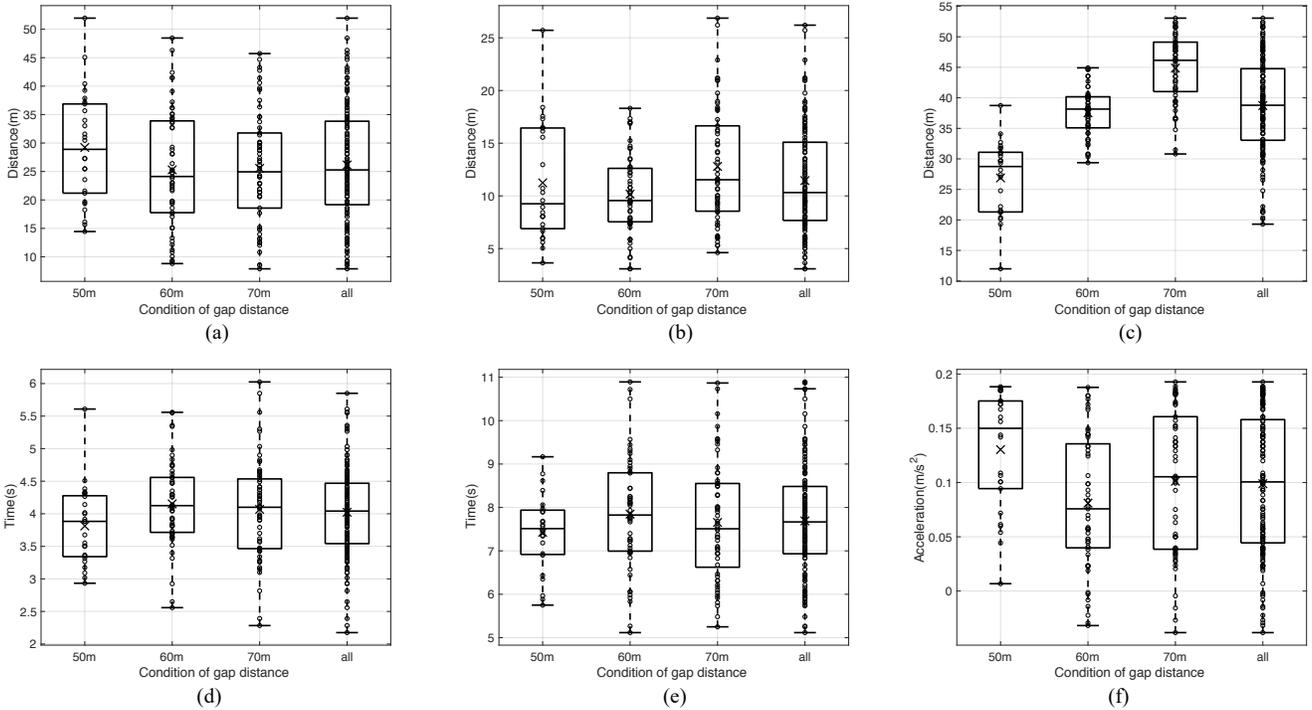

Fig. 8. Distribution of measured parameters. "50 m", "60 m", "70 m", and "all" represent that the participants made a lane change in the case of gap distance of 50 m, 60 m, 70 m, and all cases, respectively. A rectangle represents the middle 50% of a set of data. A horizontal line drawn through a rectangle corresponds to the median value of a set of data. An upper bar indicates the maximum value of a set of data, excluding outliers. A lower bar represents the minimum value of a set of data, excluding outliers. Individual results are indicated by circles. A cross symbol indicates the mean value of a set of data: (a) Measurement of $D_1$, (b) Measurement of $D_2$, (c) Measurement of $D_3$, (d) Measurement of duration between timing $t_1$ and $t_2$, (e) Measurement of duration between timing $t_1$ and $t_3$, and (f) Measurement of mean longitudinal acceleration $a_x$ during lane change maneuver.

In addition to the identified parameters of $D_1$, $D_2$, and $D_3$, during the lane change maneuver phase, the automated vehicle accelerated in the longitudinal direction with a constant value of $0.10 \text{m/s}^2$, which was the average value of $a_x$ from the identified results. The automated vehicle accelerated in the lateral direction with a constant value of $0.21 \text{m/s}^2$ from timing $t_1$ to $t_2$ and decelerated with a constant value of $0.28 \text{m/s}^2$ from timing $t_2$ to $t_3$, by referring to the equations (6) and (7). As mentioned in Section III. B, the steering wheel angle was controlled to obtain the desired lateral acceleration.

## IV. EVALUATION OF AUTOMATED LANE CHANGE SYSTEM

Another driving simulator experiment was conducted to evaluate the design of the automated lane change system with three driving styles. The subjective evaluation of the performance of the system was collected from participants. The experiment consisted of two scenarios: Scenario A and B. In Scenario A, the designed system was evaluated from the viewpoint of automated vehicle $C_0$. In Scenario B, the designed system was evaluated from the viewpoint of a surrounding vehicle.

### A. Driving Simulator Experiment

*1) Participants*

The same participants as the previous experiment, described in Section III. A. 1), were recruited again to attend this driving simulator experiment.

*2) Apparatus*

The same truck driving simulator, which was described in Section III. A. 2), was used. In this experiment, the driving speed and steering wheel angle of $C_0$ were automatedly controlled.

*3) Scenario A*

The lane change task in Scenario A is shown in Figure 9 (a). The participants acted as the driver of $C_0$ and evaluated the automated lane change system. After $C_1$ decelerated, $C_0$ attempted to make a lane change to the right lane if the gap acceptance is satisfied.

As described in Section III. C, three driving styles of automated lane change were designed. Each driving style (or driving condition) corresponded to one driving trial, as shown in Table II. The gap distance between $C_2$ and $C_3$ was kept with 45 m, 55 m, and 65 m for Trials A1, A2, and A3, respectively.

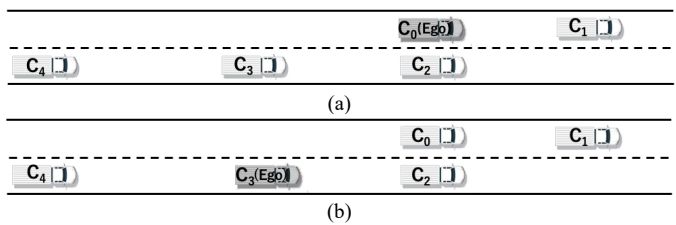

Fig. 9. Lane change scenarios in experiment for system evaluation: (a) Scenario A: $C_0$ is the ego vehicle and attempts to make a lane change, and (b) Scenario B: $C_3$ is the ego vehicle and $C_0$ attempts to make a lane change.

TABLE II
CONDITIONS OF EXPERIMENT FOR SYSTEM EVALUATION

|  | Cond. 1 (Aggressive) | Cond. 2 (Medium) | Cond. 3 (Conservative) |
|---|---|---|---|
| Scenario A | Trial A1 | Trial A2 | Trial A3 |
| Scenario B | Trial B1$_1$ Trial B1$_2$ Trial B1$_3$ | Trial B2$_1$ Trial B2$_2$ Trial B2$_3$ | Trial B3$_1$ Trial B3$_2$ Trial B3$_3$ |

As $C_0$ automatedly performed the car following and lane changing tasks, the participants were asked not to operate the steering wheel and gas/brake pedal unless necessary. The participants were requested to pay attention to the relative distances to the surrounding vehicles in order to evaluate the automated lane change performance.

To counterbalance the order of experimental conditions within the participants, 3×3 Latin Square [33] was used in the order design.

*4) Scenario B*

In Scenario B, the participants acted as the driver of $C_3$ and evaluated the automated driving performance of $C_0$ from the viewpoint of a surrounding vehicle. Thus, $C_3$ became the ego vehicle, as shown in Figure 9 (b). The participants needed to control the steering wheel to stay within the lane, and the driving speed was controlled by an adaptive cruise control system to maintain a certain distance to the lead vehicle $C_2$.

As shown in Table II, each group of trials (e. g. B1$_1$, B1$_2$, and B1$_3$) corresponded to each condition of driving style (e. g. aggressive model). The difference among the three conditions was the automated driving style of $C_0$. The difference among the three trials in each condition (e. g. B1$_1$, B1$_2$, and B1$_3$) was the gap distance between $C_2$ and $C_3$. The gap distance between $C_2$ and $C_3$ was kept with 45 m, 55 m, and 65 m for Trials B1$_1$, B1$_2$, and B1$_3$, respectively.

To counterbalance the order of experimental conditions within the participants, 3×3 Latin Square [33] was used in the order design.

*5) Procedure*

The part of the experimental procedure before practice driving was similar to the experiment for system design, as described in Section III. A. 4).

In the practice driving, the participants experienced the scenarios and familiarized themselves with the driving simulator, especially with the automated driving system in Scenario A and the adaptive cruise control system in Scenario B.

After the practice driving, the participants were allowed to rest for 5 min. After the rest, the formal experimental session started. In this session, each participant first undertook Scenario A and then Scenario B. In Scenario A, after each trial, the participants were asked to complete a questionnaire. After all the three trials in Scenario A, the participants were allowed to rest for 5 min. In Scenario B, after each trial of each condition, the participants were asked to complete a questionnaire. After all the three trials in each condition, the participants were asked to complete an overall questionnaire to evaluate each driving style. The entire experiment took approximately 120 min per participant.

*6) Measurement*

The participants' subjective evaluation of the designed automated lane change system was measured.

In Scenario A, after each driving condition, the following questions were asked:

Q1. Did you feel safe during the process of automated lane-changing? From 1 (Absolute disagree) to 5 (Absolute agree)

Q2. How do you evaluate the distance to the front car during lane-changing? From 1 (Very close) to 5 (Very far)

Q3. How do you evaluate the distance to the right front car during lane-changing? From 1 (Very close) to 5 (Very far)

Q4: How do you evaluate the distance to the right rear car during lane-changing? From 1 (Very close) to 5 (Very far)

In Scenario B, after each driving trial, the following questions were asked in the cases that the automated vehicle made a lane change and did not make a lane change:

In the case that the automated vehicle made a lane change, the following questions were asked:

Q1: Did you feel safe during the automated lane-changing of the lane-changing car? From 1 (Absolutely disagree) to 5 (Absolutely agree)

Q2: Please evaluate the distance to the lane-changing truck when the automated lane-changing system started. From 1 (Very close) to 5 (Very far)

Q3: Please evaluate the distance to the lane-changing truck after lane-changing. From 1 (Very close) to 5 (Very far)

In the case that the automated vehicle did not make a lane change, the following questions were asked:

Q4: For this trial, the automated lane-changing system estimated the scenario to be dangerous. Therefore, no lane change was made. Do you agree with its judgment? From 1 (Absolutely disagree) to 5 (Absolutely agree)

In Scenario B, after each driving condition which included three trials, the following question was asked:

Q5: Do you feel this automated lane-changing system reliable? From 1 (Unreliable) to 3 (Reliable)

*B. Data Analysis*

The subjective evaluation data were analyzed using Friedman test and Wilcoxon signed-rank test. Friedman test was used to compare three groups (e.g. aggressive, medium, and conservative models) for statistical significance. Wilcoxon signed-rank test was used to compare two groups to see which group was significantly different from each other. The Bonferroni correction was applied to control the Family-wise Error Rate. The significant level was set as 0.05.

*C. Evaluation Results*

The evaluation results on the performance of the automated driving system are presented separately for evaluation from the viewpoint of the automated vehicle (Scenario A) and from the viewpoint of a surrounding vehicle (Scenario B).

*1) From the Viewpoint of Automated Vehicle (Scenario A)*

Figure 10 shows the subjective evaluation scores for each model of driving style. Table III presents the mean value and standard deviation of evaluation scores, and the results of Friedman test and Wilcoxon signed-rank test. From the

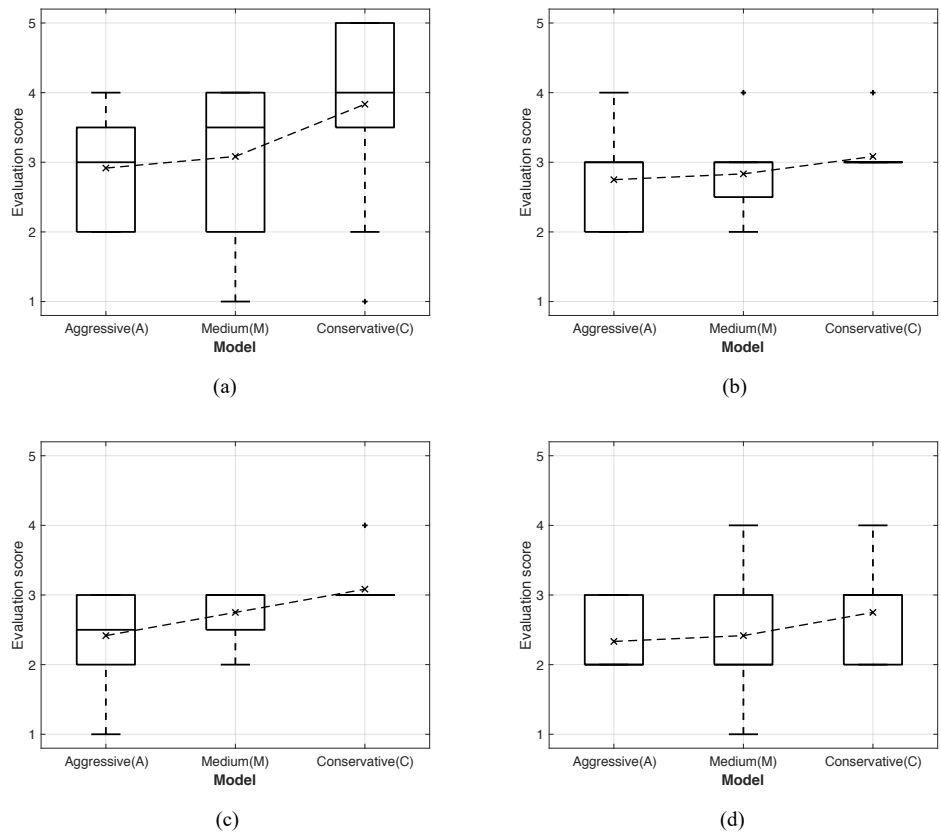

Fig. 10. Subjective evaluation scores in Scenario A. A rectangle represents the middle 50% of a set of data. A horizontal line drawn through a rectangle corresponds to the median value of a set of data. An upper bar indicates the maximum value of a set of data, excluding outliers. A lower bar represents the minimum value of a set of data, excluding outliers. Mild outliers are indicated by + symbol (calculated as 1.5–3× the interquartile range). A cross symbol indicates the mean value of a set of data: (a) Subjective evaluation on lane change safety (Q1), (b) Subjective evaluation on distance to ahead vehicle (Q2), (c) Subjective evaluation on distance to ahead vehicle of target lane (Q3), and (d) Subjective evaluation on distance to rear vehicle of target lane (Q4).

TABLE III
STATISTICAL ANALYSIS OF SUBJECTIVE EVALUATION SCORES IN SCENARIO A

|  | Aggressive M (SD) | Medium M (SD) | Conservative M (SD) | Friedman | A-M | A-C | M-C |
|---|---|---|---|---|---|---|---|
| Q1: Lane change safety | 2.92(0.79) | 3.08(1.08) | 3.83(1.27) | 0.019* | 0.531 | 0.070+ | 0.055+ |
| Q2: Distance to ahead vehicle | 2.75(0.62) | 2.83(0.58) | 3.08(0.29) | 0.197 | 1 | 0.219 | 0.375 |
| Q3: Distance to ahead vehicle of target lane | 2.42(0.67) | 2.75(0.45) | 3.08(0.29) | 0.011* | 0.25 | 0.031* | 0.25 |
| Q4: Distance to rear vehicle of target lane | 2.33(0.49) | 2.42(1.00) | 2.75(0.62) | 0.158 | 0.906 | 0.188 | 0.219 |

*$p < 0.05$, +$p < 0.1$.

Friedman test, the effect of different driving styles was significant for subjective evaluation on lane change safety and on distance to ahead vehicle of target lane.

From Wilcoxon signed-rank test in Table III and results shown in Figure 10 (a), there was a tendency that participants felt safer about the automated lane change with conservative model than with aggressive model, and with conservative model than with medium model. Moreover, for all the three driving styles, the mean value of evaluation scores was around 3, which was the median value between unsafe and safe. It indicates that, generally, the participants felt that the designed automated lane change system was acceptable on safety issues from the viewpoint of the automated vehicle.

From Wilcoxon signed-rank test in Table III and results shown in Figure 10 (c), it indicates that the distance to ahead vehicle of target lane was significantly different between the aggressive model and conservative model, and the aggressive model yielded a shorter distance. In addition, there was an increasing tendency of the distance to ahead vehicle of target lane from aggressive model to medium model and to conservative model.

As shown in Figure 10 (b), there was an increasing tendency of distance to ahead vehicle from aggressive model to medium model and to conservative model, although the difference was not significant. As shown in Figure 10 (d), there was also an increasing tendency of distance to rear vehicle of target lane from aggressive model to medium model and to conservative model, although the difference was not significant. According to the above results on the relative distance to the surrounding vehicles, it indicates that the three driving styles can be distinguished by the participants from the viewpoint of the automated vehicle.

*2) From the Viewpoint of Surrounding Vehicle (Scenario B)*

Table IV presents the mean value and standard deviation of evaluation scores, and the results of Friedman test and Wilcoxon signed-rank test. From the results, there was no significant difference among the three models of driving styles or between any two of the three models. It indicates that, in terms of the relative distance, the different driving styles were not distinguished by the participants from the viewpoint of the surrounding vehicle.

According to Table IV, the distance to the automated vehicle during lane change was evaluated by getting a mean score around or larger than 3, which was the median value between very close and very far. It indicates that the participants felt that the distance was in line with their expectation or a little far. It suggests that the automated lane change system was acceptable regarding the safe distance from the viewpoint of the surrounding vehicle.

Figure 11 shows the results of subjective evaluation scores on system reliability of automated lane change. From the results of the three driving styles, most evaluation scores fell in the area between 2 to 3 (reliable), and few participants chose 1 (unreliable). From the Friedman test, there was no significant difference in system reliability among the three driving styles. It indicates that, generally, the designed automated lane change system was reliable to the participants from the viewpoint of the surrounding vehicle.

TABLE IV
STATISTICAL ANALYSIS OF SUBJECTIVE EVALUATION SCORES IN SCENARIO B

|  | Aggressive M (SD) | Medium M (SD) | Conservative M (SD) | Friedman | A-M | A-C | M-C |
|---|---|---|---|---|---|---|---|
| Q1: Lane change safety; Comparing trial B1$_2$ and B2$_2$ | 3.42(0.79) | 3.58(0.90) | - | - | 0.688 | - | - |
| Q2: Distance to automated vehicle when lane change starts; Comparing trial B1$_2$ and B2$_2$ | 3.08(0.52) | 3.33(0.65) | - | - | 0.250 | - | - |
| Q3: Distance to automated vehicle when lane change ends; Comparing trial B1$_2$ and B2$_2$ | 2.92(0.52) | 3.00(0.74) | - | - | 1 | - | - |
| Q1: Lane change safety; Comparing trial B1$_3$, B2$_3$, and B3$_3$ | 3.58(0.90) | 3.75(0.75) | 3.75(0.75) | 0.883 | 0.750 | 0.750 | 1 |
| Q2: Distance to automated vehicle when lane change starts; Comparing trial B1$_3$, B2$_3$, and B3$_3$ | 3.50(0.52) | 3.50(0.67) | 3.50(0.67) | 0.956 | 1 | 1 | 1 |
| Q3: Distance to automated vehicle when lane change ends; Comparing trial B1$_3$, B2$_3$, and B3$_3$ | 3.33(0.49) | 3.25(0.62) | 3.17(0.67) | 0.444 | 1 | 0.625 | 1 |
| Q4: Agreement on decision of car-following rather than lane change Comparing trial B2$_1$ and B3$_1$ | - | 3.25(1.22) | 3.42(1.24) | - | - | - | 0.781 |

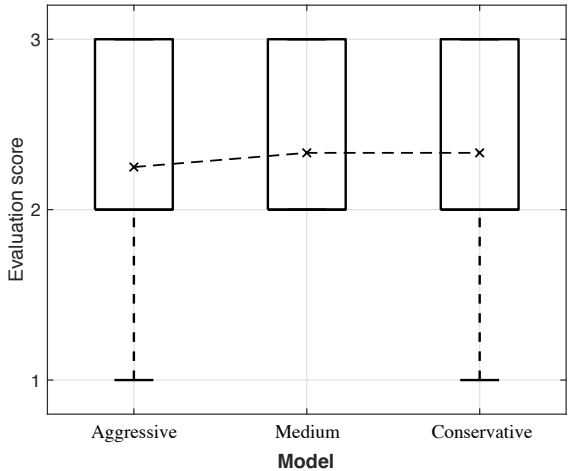

Fig. 11. Subjective evaluation scores on system reliability in Scenario B.

## V. General Discussion

### A. System Design

A discrete lane change model for truck drivers based on gap acceptance was proposed, as described in Section II. Previous research on passenger cars [32] and trucks [30] proved that the gap acceptance model was effective. Moreover, our validation study by numerical simulations demonstrated that the proposed model can yield a safe automated lane change and match the experimental results [34]. It indicates that the proposed model was capable of predicting driver lane change behavior.

It has been argued that an automated system developed for all drivers have to be conservative for the reason of safety issues to cover all types of driving styles [10]. In our study, based on the identified parameters of the driver model from 12 participants, three driving styles were classified by choosing the $25^{th}$, $50^{th}$, and $75^{th}$ percentile of the relative distances to surrounding vehicles. Although previous research classified different driving styles by adjusting the parameters in the driver model [27], our study provided a perspective on style classification by using collected experimental data from a group of drivers.

The evaluation results on the system design demonstrate that using $25^{th}$, $50^{th}$, and $75^{th}$ percentile of parameters can distinguish the different driving styles while remaining driver acceptance on the system reliability. Future studies will extend the current work by testing other percentiles of the parameters, especially in the cases of different driving scenarios.

### B. System Evaluation from Viewpoint of Automated Vehicle

The designed system was evaluated from the viewpoint of the driver in the automated vehicle, which was a common way done by other researchers [9, 23]. Our study focused on the evaluation of driver subjective feelings on the relative distance and driving safety. It has been found that a more conservative driving style would yield a longer relative distance to the surrounding vehicles and a safer feeling for the drivers [27], which is in accordance with our findings.

The statistical analysis of evaluation results indicates that the different driving styles could be distinguished by the drivers. Given that drivers prefer to adjust the level or type of automation in different situations [22], as long as the three driving styles could be distinguished, it is expected that drivers will choose a proper driving style according to the driving scenarios. For instance, experienced drivers tend to drive in a relatively aggressive way [4], and even for novel drivers, a more aggressive driving style could be preferred when he/she is in a hurry time. From the results shown in Figure 10 (a), there was a tendency that participants felt safer about the automated lane change with the conservative model than with the aggressive model. It indicates that some of the drivers would have to balance the safe feeling and lane change efficiency by choosing a proper model when they are using such a system.

In recent years, not only the safe feeling from drivers, but also the involved feeling or sense of agency have been measured to evaluate the automated driving system [19]. A higher sense of agency indicates that the driver feels more involved in the automated driving and the response time for a take-over request in a critical event can be shortened. In our follow-up study, the evaluation of driver's sense of agency will be addressed. Another improvement of our study could be to conduct physiological measurements (heart rate, eye-movement, EEG, etc.) [35, 36] to objectively evaluate driver's feelings on the automated driving system.

### C. System Evaluation from Viewpoint of Surrounding Vehicle

In addition to the system evaluation from viewpoint of the automated vehicle, this study further evaluated the system from the viewpoint of a surrounding vehicle, which is one of the first investigating from such a viewpoint. Although no significant difference in subjective evaluation was found among different driving styles as shown in Table IV, which is not in accordance with our hypothesis, this study still provides some guidance for future studies.

The driver was placed in the vehicle of the target lane behind the automated vehicle, because an unsatisfactory lane change would normally make the driver behind feel frustrated and would even cause a frontal crash [37]. It was expected that the participants can distinguish the distance to the automated vehicle when different driving styles were implemented. However, as shown in Table IV, the participants felt that the distance was in line with their expectation or a little far than expected, which indicates that the automated lane change was not risky. With regard to the result of the reliability question as shown in Figure 11, generally, the participants felt that the automated lane change system was reliable. Additionally, all three driving styles yielded a similar level of reliability, which is in accordance with the results of self-evaluation on the relative distance.

By looking at the results from viewpoint of the automated vehicle as shown in Table III, with regard to the distance to the rear vehicle of target lane, there was no significant difference among the three driving styles, which is in accordance with the result from viewpoint of the surrounding vehicle. One explanation could be that the current lane change scenario was relatively simple because the driving speed of surrounding vehicles was constant; thus, when the participants felt that the automated system is reliable, their attention on the relative distance between the automated vehicle and rear vehicle was reduced. In future work, a more complicated lane change scenario with the variation of driving speed will be designed.

VI. CONCLUSION

A truck driving simulator experiment with 12 participants was conducted to analyze and model truck driver behavior, by which an automated lane change system with different driving styles (aggressive, medium, and conservative) was designed. Another truck driving simulator experiment with the same 12 participants was conducted to evaluate the proposed automated driving system in terms of driver experience and acceptance.

The evaluation results indicate that the different lane change styles could be distinguished by the drivers from the viewpoint of the automated vehicle; meanwhile, the three styles were overall evaluated as acceptable on safety issues and reliable by the drivers from the viewpoints of both the automated vehicle and the surrounding vehicle. This study provides insight into designing an automated driving system with different driving styles and the findings can be applied to commercial automated trucks.

A limitation of the experiment was that the lane change scenario was relatively simple and future work will introduce a more complicated scenario. In addition, only subjective evaluation of the designed system was conducted in the current study and future studies will address this limitation by including objective evaluation.